# THE IMPACT OF DIFFERENT BACKBONE ARCHITECTURE ON AUTONOMOUS VEHICL DATASET

**Ning Ding, Azim Eskandarian**

Virginia Polytechnic Institute and State University, Blacksburg, VA, USA

**ABSTRACT**

*Object detection is a crucial component of autonomous driving, and many detection applications have been developed to address this task. These applications often rely on backbone architectures, which extract representation features from inputs to perform the object detection task. The quality of the features extracted by the backbone architecture can have a significant impact on the overall detection performance. Many researchers have focused on developing new and improved backbone architectures to enhance the efficiency and accuracy of object detection applications. While these backbone architectures have shown state-of-the-art performance on generic object detection datasets like MS-COCO and PASCAL-VOC, evaluating their performance under an autonomous driving environment has not been previously explored. To address this, our study evaluates three well-known autonomous vehicle datasets, namely KITTI, NuScenes, and BDD, to compare the performance of different backbone architectures on object detection tasks.*

## 1. INTRODUCTION

Object detection (OD) is a crucial component of autonomous driving. Many current OD applications for autonomous driving come from the research of Computer Vision's contribution on 2D OD [1]. Since the AlexNet [2], deep-learning based approaches has been the dominant trend in this area compared with traditional approaches [3-5]. In deep deep-learning based approaches, it is common to employ an architecture, known as backbone, consisting of multiple layers to extract features from input images, and produce a condensed and meaningful representation. This feature representation can then be utilized for downstream object detection task. And the quality of produced feature representation can greatly impact the accuracy and efficiency of object detection. Therefore, significant research has been conducted on backbones in deep learning, resulting in several effective architectures that have been widely adopted in popular object detection models. Examples of well-known backbones that have demonstrated state-of-the-art performance in object detection tasks include ResNet [6], Darknet [10], Res2Net [8], and Swin Transformer (Swin-T) [9].

While all of the aforementioned backbone architectures have been proven effective in several object detection models, researchers who propose these architectures typically evaluate them on generic datasets such as COCO and ImageNet. Few studies have focused on evaluating the performance of these architectures on autonomous vehicle datasets specifically, which have their own unique characteristics. Characteristics that pose challenges for object detection include the presence of objects in images that are too small for human eyes to discern, categories (such as cars, trucks, and vans) whose shapes and sizes are very similar, and low-quality data resulting from images captured in poor weather conditions. As a result, evaluating the performance of object detection models on autonomous vehicle datasets is important to ensure their applicability in real-world scenarios.

Thus, our study concentrates on autonomous vehicles 2D datasets and aims to evaluate the performance of four aforementioned backbone architectures using the same 2D object detection model. We use the Deformable DETR as the only object detection model to evaluate the performance of different backbones, because previous research [11] has shown that Deformable DETR is able to effectively utilize the features from different backbones to achieve better object detection results compared to other detector model on the COCO dataset.

The contribution of this paper includes:
- Four different backbone architectures are evaluated.
- Three different autonomous vehicle datasets (KITTI [15], nuScenes [16], and BDD [17]) are investigated in this study.
- The performance of a transformer-based detector (Deformable-DETR) with different backbones on autonomous vehicle datasets is presented in this study. To the best of our knowledge, this is the first time such an analysis has been conducted.

## 2. METHODS

In this section, we begin by providing a detailed description of the four backbones examined in this study. Following that, we introduce Deformable-DETR, which is utilized as the detector in our research. Finally, we present the evaluation metrics employed.

### 2.1 ResNet

ResNet is a type of Convolutional Neural Network (CNN), whose main innovation is building a residual learning block in the network, defined as:

$$y = \mathcal{F}(x, \{W_i\}) + W_s x \quad (1)$$

Where, $x$ is the input of layers, $y$ is the output of layers, $\mathcal{F}(x, \{W_i\})$ represents fully connected layers, $W_s$ is the square matrix used to match the dimension. By utilizing this residual learning technique, ResNet is able to solve the issue of vanishing gradients in deep neural networks. The residual connections within these models facilitate the flow of gradients through deeper layers of the network, which leads to improved performance and faster convergence.

We choose ResNet-50 and ResNet-101 based on ResNet in our study. These two architectures comprise of 50 and 101 convolutional layers respectively. While ResNet-101 is generally regarded as an improvement over ResNet-50 [6], the actual performance gain can vary depending on the task and dataset being considered. This study will show if the ResNet is suitable for autonomous driving datasets and assess whether increasing the number of layers in ResNet can result in significant improvements in performance.

**2.2 DarkNet**

The Darknet [10] architecture is composed of several convolutional layers, followed by a global average pooling layer, and fully connected layers for prediction. The Darknet-53 [7] architecture used in our study is based on the Darknet and includes 53 convolutional layers with residual connections. These additional residual connections, which are not included in original Darknet, allow for easier training of deep neural networks.

Darknet-53 has been introduced as an alternative architecture to ResNet for object detection and is computationally less expensive. In our study, we aim to investigate whether Darknet-53 with simpler architecture can catch up to ResNet when handling autonomous vehicle datasets.

**2.3 Res2Net**

Res2Net is an extension of ResNet that introduces a novel residual block with multiple hierarchical residual connections. Within each residual block, the output of the preceding layer is combined with feature maps of different scales in parallel. This design enables the network to capture multi-scale features and enhances its ability to detect objects with various sizes and shapes.

Our study employs the Res2Net-101 architecture, which consists of 101 layers and has demonstrated better performance on generic datasets such as COCO [8]. The objective of our investigation is to determine whether Res2Net-101 can also enhance the object detection performance on autonomous vehicle datasets when compared with other CNN-based backbone architectures.

**2.4 Swin Transformer (Swin-T)**

Following the success of attention mechanisms in natural language processing tasks, researchers have also applied this technique to object detection tasks and demonstrated its effectiveness [12].

Swin-T is a symbolized architecture applying attention mechanism. The Swin block is the fundamental block of the Swin-T, using a self-attention mechanism with shifted windows to capture features, resulting in the produced feature representation including spatial differences between different regions of an image. By processing the image in a hierarchical manner with multiple stages of Swin blocks, the Swin-T is able to effectively capture both local and global features from the input image.

We choose Swin-T as one of the selected backbones for investigation, which will help us evaluate the performance of an transformer-based backbone in tasks related to autonomous vehicles.

**2.5 Deformable-DETR**

DETR [13] is the first object detection model employing a transformer-based architecture different from traditional CNN-based approaches. This architecture encodes the entire feature map generated by the backbone network, leveraging a self-attention mechanism to capture global contextual information. This approach enables DETR to make informed decisions regarding object detection, and it has achieved state-of-the-art performance on COCO dataset. Despite its breakthrough performance, DETR faces several challenges. One of these is that it can be difficult to train, particularly on large-scale dataset. Another challenge is its difficulty in detecting small objects.

Deformable-DETR was introduced later as an enhanced version of DETR to address these challenges. It does this by incorporating deformable attention modules to modify its attention mechanism. To prevent the model from becoming too complex and to aid convergence, Deformable-DETR employs a limited number of keys for each query. Furthermore, it employs multi-scale feature maps, allowing it to excel at detecting smaller objects.

**2.6 Evaluation Metrics**

Our evaluation metrics come from COCO's evaluation method, which are based on Precision (2) and Recall (3). High Precision means that the model makes fewer false positive predictions, and high Recall means that the model captures a larger proportion of true positive instances.

$$Precision = \frac{TP}{TP+FP} \quad (2)$$

$$Recall = \frac{TP}{TP+FN} \quad (3)$$

Where, TP represents true positive predictions, FP represents false positive predictions, FN represents false negative predictions.

Precision and Recall can be used to calculate Average Precision (AP), mean Average Precision (mAP), Average Recall (AR) and mean Average Recall (mAR). AP is the mean precision value over varying detection threshold, mAP is the average value of all APs over all categories. AR represents the average recall value over all detection thresholds, while mAR is the average of ARs computed over all categories. Intersection over Union (IoU) is used as detection threshold metric in the study. To calculate the IoU value, the intersection between the predicted and ground truth regions is divided by their union. A threshold value of IoU means that the prediction is considered a true positive if the IoU value between the predicted bounding box and ground truth is greater than that threshold value. In the study, 10 IoU thresholds are used to calculate AP, mAP, AR and mAR. They are 0.5, 0.55, 0.6, 0.65, 0.7, 0.75, 0.8, 0.85, 0.9 and 0.95. We also report mAP at IoU=0.5 and 0.75 to provide a more complete picture of detection performance. To investigate the impact of instance size on detection performance, we group all instances based on their areas, which are measured in terms of the number of pixels. Following COCO's evaluation method, instances with an area less than $32^2$ are categorized as small, those with an area between $32^2$ and $96^2$ were categorized as medium, and those with an area greater than $96^2$ are categorized as large. We present the mAP and mAR for each of these three groups.

To analyze the influence of different backbones on the inference speed, we measure the total time encompassing data preprocessing, detection model inference, and image post-processing. The metric is presented as frames per second (FPS).

In summary, Table 1 displays all evaluation metrics we use.

**Table 1:** EVALUATION METRICS

| Metric | Explanation |
|---|---|
| AP | AP for each category |
| mAP | Mean AP over all categories |
| $mAP_{0.5}$ | mAP at IoU=0.5 |
| $mAP_{0.75}$ | mAP at IoU=0.75 |
| $mAP_S$ | mAP for instances: area<$32^2$ |
| $mAP_M$ | mAP for instances: $32^2$<area<$96^2$ |
| $mAP_L$ | mAP for instances: area>$96^2$ |
| mAR | Mean AR over all categories |
| $mAR_S$ | mAR for instances: area<$32^2$ |
| $mAR_M$ | mAR for instances: $32^2$<area<$96^2$ |
| $mAR_L$ | mAR for instances: area>$96^2$ |
| FPS | Inference speed |

## 3. Experiments and Results

### 3.1 Dataset

#### 3.1.1 KITTI

**Table 2:** INSTANCES OF EACH CATEGORY IN KITTI

|  | train | validation |
|---|---|---|
| Car | 23,174 | 5,567 |
| Van | 2,344 | 568 |
| Truck | 900 | 194 |
| Pedestrian | 3,630 | 856 |
| Person Sitting | 172 | 50 |
| Cyclist | 1,291 | 336 |
| Tram | 410 | 101 |
| Misc | 802 | 170 |

The KITTI Vision Benchmark Suite is a popular dataset used in autonomous vehicle research [14], featuring a dedicated section for object detection. KITTI dataset includes stereoscopic images; however, for the purpose of ensuring image uniqueness, we exclusively utilized images from the left camera perspective in our study. This section contains 7,481 labeled images, comprising of 80,256 labeled objects from diverse scenes. The labeled objects are classified into eight categories: car, van, truck, pedestrian, person sitting, cyclist, tram, and misc (e.g. trailers). To create the training and validation sets, we randomly selected 5,984 and 1,497 images respectively from labeled images. Table 2 presents the instances of each category in the training and test sets.

#### 3.1.2 nuScenes

**Table 3:** INSTANCES OF EACH CATEGORY IN NUSCENES

|  | train | validation |
|---|---|---|
| Pedestrian | 135,870 | 32,710 |
| Barrier | 70,112 | 18,433 |
| Road object | 5,491 | 1,355 |
| Traffic cone | 69,016 | 18,587 |
| Bike | 16,169 | 3,955 |
| Bus | 6,741 | 1,885 |
| Car | 202,947 | 47,322 |
| Construction Vehicle | 4,768 | 1,303 |
| Motor | 13,682 | 3,097 |
| Trailer | 3,285 | 486 |
| Truck | 29,456 | 6,858 |

The NuScenes dataset is a relatively new and more challenging dataset compared to KITTI. Its object detection section, called nuImages, includes 60,000 labeled images for training and 16,000 labeled images for validation, captured from 1,000 different scenes. The dataset encompasses 91,750 labeled objects belonging to fifteen categories, such as pedestrian, barrier, debris, pushable-pullable objects, traffic cone, bicycle rack, bicycle, bus, car, construction vehicle, emergency vehicle, motor, trailer, and truck. To simplify the classification, we merged some similar categories into one group. For instance, we combined *bicycle* and *bicycle rack* into a single class called *bike* as we did not consider explicitly differentiating based on the motion of the bicycle. *Debris* and *pushable-pullable objects* were merged into *road object*. *Emergency vehicle* is included into *car*. Moreover, we removed the category *animal* due to its infrequent occurrence in driving situations. Table 3 lists the instances of each category in our study.

#### 3.1.3 BDD

**Table 4:** INSTANCES OF EACH CATEGORY IN BDD

|  | train | validation |
|---|---|---|
| Traffic sign | 175,724 | 34,908 |
| Traffic light | 159,000 | 26,885 |
| Pedestrian | 87,174 | 13,262 |
| Car | 443,160 | 102,506 |
| Motor | 3,002 | 452 |
| Bus | 11,672 | 1,597 |
| Bike | 7,210 | 1,007 |
| Truck | 28,149 | 4,245 |
| Rider | 4,517 | 649 |

The BDD dataset comprises images captured in different weather conditions and at varying times of the day. With over 1,840,000 labeled objects, the dataset includes a training set of 70,000 labeled images and a validation set of 10,000 labeled images. The dataset features ten labeled categories: traffic sign, traffic light, pedestrian, car, motor, bus, bike, truck, train, and rider. Since the category *train* contains only 179 instances, much lower than the other classes, we exclude it from our analysis. Table 4 summarizes the instances for each category.

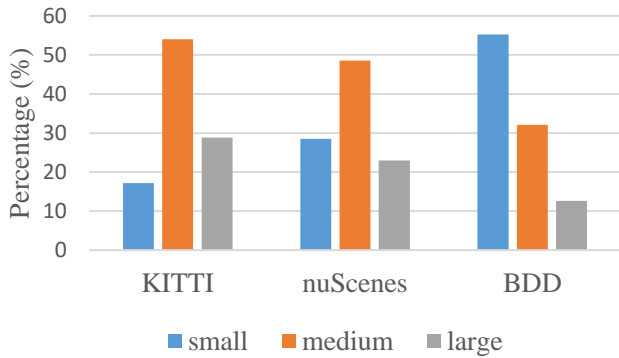

**FIGURE 1:** PERCENTAGE OF SMALL, MEDIUM AND LARGE INSTANCES IN EACH DATASET

Figure 1 illustrates the percentages of small, medium, and large instances in three datasets. It reveals that the proportion of small instances is considerably higher in nuScenes and BDD compared to KITTI, rendering these two datasets more challenging. Notably, in the BDD dataset, over 50% of the instances belong to the small category, which highlights its exceptional level of difficulty.

### 3.2 Implementation Details

The Deformable-DETR model with different backbone is built using PyTorch deep learning platform. The training and validation of the model are carried out on a NVIDIA A100-80GB GPU, with a batch size of 16 and 1, respectively. During training, the hyperparameters are set according to [11], and the model is trained for 50 epochs using the AdamW [18] optimizer. The initial learning rate is set to 2e-4 and is decayed from the 40th epoch by a factor of 0.01. Additionally, the learning rates of the linear projections are reduced by a factor of 0.1.

### 3.3 Results and Analysis

#### 3.3.1 KITTI

Based on the Table 5, increasing the number of layers in a ResNet backbone network generally leads to better performance, but this improvement is not as significant as switching to a more advanced backbone network such as Res2Net-101 or Swin Transfer. Darknet-53, which is a lightweight network, may have inferior performance compared to other networks. However, its performance is more comparable to ResNet-50 when evaluating the $mAP_{0.5}$ metric. In fact, the difference in $mAP_{0.5}$ between Darknet-53 and ResNet-50 is not as significant as the difference in $mAP_{0.75}$.

Res2Net-101 outperforms all other CNN-based backbone architectures tested in the experiment. This is primarily attributed to its superior ability to detect small objects, which is the best among all the backbones evaluated.

While Res2Net-101 outperforms Swin-T in detecting small objects, Swin-T remains the best-performing backbone overall in most metrics. As shown in Figure 1, the percentage of training instances for small objects is less than 20%, which is much lower than that of medium and large objects. This suggests that the limited exposure to small objects during training may be a reason why Swin-T struggles with detecting them. It is important to note that Swin-T exhibits the slowest inference speed among the four backbones. Therefore, a trade-off needs to be considered between performance and inference speed.

**Table 5:** DETECTION PERFORMANCE OVER ALL CATEGORIES ON KITTI

|  | mAP | $mAP_{0.5}$ | $mAP_{0.75}$ | $mAP_S$ | $mAP_M$ | $mAP_L$ | mAR | $mAR_S$ | $mAR_M$ | $mAR_L$ | FPS |
|---|---|---|---|---|---|---|---|---|---|---|---|
| Resnet-50 | 0.641 | 0.883 | 0.732 | 0.641 | 0.651 | 0.675 | 0.731 | 0.705 | 0.733 | 0.772 | **6.2** |
| Resnet-101 | 0.651 | 0.900 | 0.732 | 0.625 | 0.668 | 0.688 | 0.741 | 0.699 | 0.746 | 0.779 | 6.0 |
| Darknet-53 | 0.618 | 0.869 | 0.694 | 0.601 | 0.640 | 0.637 | 0.705 | 0.668 | 0.717 | 0.736 | 6.0 |
| Res2Net-101 | 0.678 | 0.909 | 0.774 | **0.690** | 0.689 | 0.704 | 0.755 | **0.744** | 0.759 | 0.785 | 4.7 |
| Swin-T | **0.697** | **0.926** | **0.800** | 0.682 | **0.704** | **0.735** | **0.777** | 0.740 | **0.778** | **0.817** | 3.9 |

**Table 6:** MAP OF EACH CATEGORY IN KITTI

|  | Car | Truck | Van | Pedestrian | Person Sitting | Tram | Cyclist | Misc |
|---|---|---|---|---|---|---|---|---|
| Resnet-50 | 0.771 | 0.830 | 0.761 | 0.444 | 0.405 | 0.683 | 0.582 | 0.653 |
| Resnet-101 | 0.781 | 0.832 | 0.762 | 0.446 | 0.441 | 0.685 | 0.589 | 0.671 |
| Darknet-53 | 0.764 | 0.817 | 0.741 | 0.412 | 0.382 | 0.686 | 0.546 | 0.598 |
| Res2Net-101 | 0.796 | 0.848 | 0.771 | 0.457 | 0.495 | 0.713 | **0.641** | 0.701 |
| Swin-T | **0.811** | **0.863** | **0.784** | **0.507** | **0.524** | **0.758** | 0.631 | **0.703** |

What is interesting in Table 6 is that the performance on Person Sitting is not so bad even if the number of its training instances is only 172. This could be attributed to the fact that the features of Person Sitting are similar to Pedestrian, and a larger number of training instances of Pedestrian helps the backbone to extract features that are also relevant to Person Sitting. Therefore, the detector can leverage the features learned from the Pedestrian category to improve the detection performance on the Person Sitting, even with fewer training instances.

### 3.3.2 nuScenes

The nuScenes dataset provides more instances for training and validation. As shown in Table 7, the performance gap between Resnet-50, Resnet-101, and Darknet-53 narrows as the number of training instances increases. Compared with KITTI dataset, Darknet-53 even outperforms Resnet-50 and Resnet-101 in detecting small objects.

Res2Net-101 remains the top-performing CNN-based backbone architecture. However, unlike the KITTI dataset, training on more instances enables Swin Transfer to focus more on relevant informative regions, leading to better performance in detecting small objects than Res2Net-101.

Table 8 shows that Trailer, Construction Vehicles, and Road Object are the worst performing categories, possibly because their number of training instances is not on the same order of magnitude as the other categories. Interestingly, similar to the category Person Sitting in the KITTI dataset, the performance on Bus is not as low as Trailer, Construction Vehicles, and Road Object, despite having a smaller number of training instances. This could also be due to that Bus shares features with Car, which has a larger number of training instances.

**Table 7:** DETECTION PERFORMANCE OVER ALL CATEGORIES ON NUSCENES

|            | mAP   | mAP$_{0.5}$ | mAP$_{0.75}$ | mAP$_S$ | mAP$_M$ | mAP$_L$ | mAR   | mAR$_S$ | mAR$_M$ | mAR$_L$ | FPS |
|------------|-------|-------------|--------------|---------|---------|---------|-------|---------|---------|---------|-----|
| Resnet-50  | 0.400 | 0.659       | 0.416        | 0.197   | 0.383   | 0.551   | 0.567 | 0.373   | 0.552   | 0.703   | **3.4** |
| Resnet-101 | 0.402 | 0.663       | 0.418        | 0.191   | 0.385   | 0.560   | 0.567 | 0.364   | 0.551   | 0.709   | 3.3 |
| Darknet-53 | 0.399 | 0.657       | 0.416        | 0.208   | 0.386   | 0.542   | 0.567 | 0.388   | 0.550   | 0.704   | 3.3 |
| Res2Net-101| 0.436 | 0.693       | 0.465        | 0.226   | 0.422   | 0.578   | 0.593 | 0.396   | 0.577   | 0.721   | 2.8 |
| Swin-T     | **0.459** | **0.733** | **0.491**   | **0.251** | **0.436** | **0.599** | **0.617** | **0.449** | **0.605** | **0.743** | 2.2 |

**Table 8:** MAP OF EACH CATEGORY IN NUSCENES

|            | Pedestrian | Motor | Bike  | Traffic Cone | Car   | Trailer | Truck | Road Object | Bus   | Construction Vehicle | Barrier |
|------------|-----------|-------|-------|--------------|-------|---------|-------|-------------|-------|---------------------|---------|
| Resnet-50  | 0.419     | 0.527 | 0.481 | 0.464        | 0.611 | 0.153   | 0.450 | 0.138       | 0.463 | 0.258               | 0.441   |
| Resnet-101 | 0.421     | 0.519 | 0.491 | 0.457        | 0.609 | 0.182   | 0.440 | 0.146       | 0.460 | 0.267               | 0.434   |
| Darknet-53 | 0.434     | 0.522 | 0.480 | 0.467        | 0.624 | 0.150   | 0.454 | 0.128       | 0.449 | 0.241               | 0.445   |
| Res2Net-101| 0.466     | 0.544 | 0.509 | 0.506        | 0.630 | 0.195   | 0.501 | 0.168       | 0.510 | 0.301               | 0.471   |
| Swin-T     | **0.487** | **0.564** | **0.530** | **0.511** | **0.642** | **0.245** | **0.518** | **0.211** | **0.533** | **0.329** | **0.479** |

### 3.3.3 BDD

As shown in Figure I, BDD dataset has a higher proportion of small objects, which can make detection more challenging. This is reflected in the performance results presented in Table 9, where the mAP$_{0.75}$ metric is reduced by approximately 55% compared to mAP$_{0.5}$. In contrast, the decrease is around 20% in KITTI dataset and 36% in nuScenes dataset.

As observed in the nuScenes dataset, a sufficiently large number of training instances can narrow the performance gap between Resnet-50, Resnet-101, and Darknet-53. Similarly, increasing the number of training instances for small objects can also narrow the performance gap between Res2Net-101 and Swin Transfer for detecting small objects.

The Motor, Bike, and Rider categories in the BDD dataset have the lowest number of training instances, and they are dissimilar to categories with a large number of training instances. This means that they are not able to utilize the features learned from other categories to improve their detection performance. Under such circumstances, Swin-T performs better than CNN-based backbones on these three categories, despite their limited number of training instances.

**Table 9:** DETECTION PERFORMANCE OVER ALL CATEGORIES ON BDD

|            | mAP   | mAP$_{0.5}$ | mAP$_{0.75}$ | mAP$_S$ | mAP$_M$ | mAP$_L$ | mAR   | mAR$_S$ | mAR$_M$ | mAR$_L$ | FPS |
|------------|-------|-------------|--------------|---------|---------|---------|-------|---------|---------|---------|-----|
| Resnet-50  | 0.296 | 0.579       | 0.262        | 0.136   | 0.345   | 0.537   | 0.449 | 0.276   | 0.516   | 0.661   | **3.7** |
| Resnet-101 | 0.297 | 0.583       | 0.266        | 0.129   | 0.349   | 0.538   | 0.442 | 0.263   | 0.509   | 0.660   | 3.5 |
| Darknet-53 | 0.290 | 0.576       | 0.252        | 0.132   | 0.344   | 0.499   | 0.446 | 0.277   | 0.515   | 0.628   | 3.5 |
| Res2Net-101| 0.308 | 0.606       | 0.268        | 0.146   | 0.357   | 0.537   | 0.461 | 0.290   | 0.523   | 0.658   | 2.9 |
| Swin-T     | **0.328** | **0.630** | **0.295**   | **0.148** | **0.372** | **0.580** | **0.469** | **0.297** | **0.529** | **0.688** | 2.2 |

Table 10: MAP OF EACH CATEGORY IN BDD

| | Motor | Traffic Light | Truck | Bike | Bus | Car | Traffic Sign | Pedestrian | Rider |
|---|---|---|---|---|---|---|---|---|---|
| Resnet-50 | 0.178 | 0.211 | 0.382 | 0.202 | 0.401 | 0.423 | 0.336 | 0.310 | 0.218 |
| Resnet-101 | 0.175 | 0.225 | 0.390 | 0.204 | 0.406 | 0.427 | 0.338 | 0.306 | 0.203 |
| Darknet-53 | 0.152 | 0.219 | 0.379 | 0.195 | 0.406 | 0.429 | 0.334 | 0.304 | 0.190 |
| Res2Net-101 | 0.193 | 0.228 | 0.397 | 0.217 | 0.416 | **0.442** | 0.347 | **0.327** | 0.205 |
| Swin-T | **0.229** | **0.230** | **0.431** | **0.236** | **0.472** | 0.441 | **0.364** | 0.319 | **0.229** |

## 4. CONCLUSION

Our study examines the effect of various backbones on detection performance of Deformable DETR using an autonomous vehicle datasets. We discover that (1) Increasing the number of training instances can reduce the performance gap between Resnet with different convolutional layers and between Resnet and Darknet. (2) Swin-T, which is a Transformer-based backbone, has been found to outperform CNN-based backbones on almost all evaluation metrics. However, it may struggle with detecting small objects when the number of training instances is limited. It also exhibits the slowest inference speed among tested backbones. (3) Res2Net can achieve comparable performance to Swin-T in detecting small objects, provided that the proportion of small objects in the training instances is very high. (4) Categories with ample training instances can enhance the detection performance of similar categories with limited training instances. Our finding could not be explored and investigated if such study is only conducted on generic datasets. We hope that our study can provide a useful reference for researchers who wish to utilize relevant backbones under autonomous driving condition in their future research.

## REFERENCES

[1] Eskandarian A, Wu C, Sun C. Research advances and challenges of autonomous and connected ground vehicles. IEEE Transactions on Intelligent Transportation Systems. 2019 Dec 18;22(2):683-711.

[2] Krizhevsky A, Sutskever I, Hinton GE. Imagenet classification with deep convolutional neural networks. Communications of the ACM. 2017 May 24;60(6):84-90.

[3] Ghorai P, Eskandarian A, Kim YK, Mehr G. State estimation and motion prediction of vehicles and vulnerable road users for cooperative autonomous driving: A survey. IEEE Transactions on Intelligent Transportation Systems. 2022 Apr 13;23(10):16983-7002.

[4] Lowe DG. Distinctive image features from scale-invariant keypoints. International journal of computer vision. 2004 Nov;60:91-110.

[5] Zhang C, Eskandarian A. A quality index metric and method for online self-assessment of autonomous vehicles sensory perception. arXiv preprint arXiv:2203.02588. 2022 Mar 4.

[6] He K, Zhang X, Ren S, Sun J. Deep residual learning for image recognition. InProceedings of the IEEE conference on computer vision and pattern recognition 2016 (pp. 770-778).

[7] Redmon J, Farhadi A. Yolov3: An incremental improvement. arXiv preprint arXiv:1804.02767. 2018 Apr 8.

[8] Gao SH, Cheng MM, Zhao K, Zhang XY, Yang MH, Torr P. Res2net: A new multi-scale backbone architecture. IEEE transactions on pattern analysis and machine intelligence. 2019 Aug 30;43(2):652-62.

[9] Liu Z, Lin Y, Cao Y, Hu H, Wei Y, Zhang Z, Lin S, Guo B. Swin transformer: Hierarchical vision transformer using shifted windows. InProceedings of the IEEE/CVF international conference on computer vision 2021 (pp. 10012-10022).

[10] Redmon J, Divvala S, Girshick R, Farhadi A. You only look once: Unified, real-time object detection. InProceedings of the IEEE conference on computer vision and pattern recognition 2016 (pp. 779-788).

[11] Zhu X, Su W, Lu L, Li B, Wang X, Dai J. Deformable detr: Deformable transformers for end-to-end object detection. arXiv preprint arXiv:2010.04159. 2020 Oct 8.

[12] Zhang C, Eskandarian A, Du X. Attention-based neural network for driving environment complexity perception. In2021 IEEE International Intelligent Transportation Systems Conference (ITSC) 2021 Sep 19 (pp. 2781-2787). IEEE.

[13] Carion N, Massa F, Synnaeve G, Usunier N, Kirillov A, Zagoruyko S. End-to-end object detection with transformers. InComputer Vision–ECCV 2020: 16th European Conference, Glasgow, UK, August 23–28, 2020, Proceedings, Part I 16 2020 (pp. 213-229). Springer International Publishing.

[14] Zhao ZQ, Zheng P, Xu ST, Wu X. Object detection with deep learning: A review. IEEE transactions on neural networks and learning systems. 2019 Jan 27;30(11):3212-32.

[15] Geiger A, Lenz P, Stiller C, Urtasun R. Vision meets robotics: The KITTI dataset. The International Journal of Robotics Research. Int J Rob Res. 2013:1-6.

[16] Caesar H, Bankiti V, Lang AH, Vora S, Liong VE, Xu Q, Krishnan A, Pan Y, Baldan G, Beijbom O. nuscenes: A multimodal dataset for autonomous driving. InProceedings of the IEEE/CVF conference on computer vision and pattern recognition 2020 (pp. 11621-11631).

[17] Yu F, Chen H, Wang X, Xian W, Chen Y, Liu F, Madhavan V, Darrell T. Bdd100k: A diverse driving dataset for heterogeneous multitask learning. InProceedings of the IEEE/CVF conference on computer vision and pattern recognition 2020 (pp. 2636-2645).

[18] Loshchilov I, Hutter F. Decoupled weight decay regularization. arXiv preprint arXiv:1711.05101. 2017 Nov 14